\documentclass[conference]{IEEEtran}
\IEEEoverridecommandlockouts
\usepackage{cite}
\usepackage{amsmath,amssymb,amsfonts}
\usepackage{algorithmic}
\usepackage{graphicx}
\usepackage{textcomp}
\usepackage{xcolor}
\usepackage{multirow}
\usepackage{float}

\def\BibTeX{{\rm B\kern-.05em{\sc i\kern-.025em b}\kern-.08em
    T\kern-.1667em\lower.7ex\hbox{E}\kern-.125emX}}
    
\begin{document}

\title{Advanced Segmentation of Diabetic Retinopathy Lesions Using DeepLabv3+\\

}

 \author{\IEEEauthorblockN{Meher Boulaabi}
 \IEEEauthorblockA{\textit{University of Monastir, FSM } \\
 \textit{University of Tunis, ENSIT-LaTICE, Tunis, Tunisia}\\
 boulaabimeher.54@gmail.com}
 \and

 \IEEEauthorblockN{Takwa Ben Aïcha Gader} 
 \IEEEauthorblockA{\textit{University of Tunis} \\
 \textit{ENSIT - LaTICE Laboratory, Tunis, Tunisia }\\
 takwa.ben.aichaa@gmail.com}
 \and

 \IEEEauthorblockN{Afef Kacem Echi}
 \IEEEauthorblockA{\textit{University of Tunis} \\
 \textit{ENSIT - LaTICE Laboratory, Tunis, Tunisia }\\
 afef.kacem@ensit-u.tunis.tn}
 \and

 \IEEEauthorblockN{Sameh Mbarek}
 \IEEEauthorblockA{\textit{Associate Professor - Faculty of Medicine of Monastir, Tunisia} \\
 \textit{Department of Ophthalmology, Taher Sfar Mahdia Hospital}\\
 Monastir, Tunisia \\
 samahmbarek@gmail.com}

 }

\maketitle

\begin{abstract} 

To improve the segmentation of diabetic retinopathy lesions (microaneurysms, hemorrhages, exudates, and soft exudates), we implemented a binary segmentation method specific to each type of lesion. As post-segmentation, we combined the individual model outputs into a single image to better analyze the lesion types. This approach facilitated parameter optimization and improved accuracy, effectively overcoming challenges related to dataset limitations and annotation complexity. Specific preprocessing steps included cropping and applying contrast-limited adaptive histogram equalization to the L channel of the LAB image. Additionally, we employed targeted data augmentation techniques to further refine the model's efficacy. Our methodology utilized the DeepLabv3+ model, achieving a segmentation accuracy of 99\%. These findings highlight the efficacy of innovative strategies in advancing medical image analysis, particularly in the precise segmentation of diabetic retinopathy lesions. The IDRID dataset was utilized to validate and demonstrate the robustness of our approach. 

\end{abstract}


\section{Introduction}

Diabetic retinopathy (DR) is a serious ocular condition impacting individuals with diabetes. Elevated blood sugar levels lead to detrimental effects on the blood vessels within the retina, the light-sensitive tissue situated at the posterior part of the eye \cite{b1}. DR is one of the leading causes of blindness among adults in developed countries. As the duration of diabetes increases, so does the risk of developing DR. Consistently high blood sugar levels, hypertension, and high cholesterol levels can also contribute to the development of DR \cite{b1}. Pregnant women with diabetes are also potentially at increased risk. Effective diabetes management is crucial to preventing or delaying the progression of DR\cite{b2}. This includes monitoring blood sugar levels regularly to stay within the recommended target range, controlling blood pressure through lifestyle changes and medication, managing cholesterol levels with a balanced diet, exercise, and medications, and undergoing an annual comprehensive dilated eye exam.
Proper management of diabetic retinopathy (DR) requires a clear understanding of its severity and impact on vision. The treatment options for DR vary depending on these factors and may include laser therapy, anti-VEGF injections, or vitrectomy surgery \cite{b3}. Regular eye exams are essential for early detection of DR when treatment is most effective. Early detection and timely intervention can prevent the vision loss and blindness associated with DR.

\begin{figure}[htbp]
    \centerline{\includegraphics[width = 0.4\textwidth]{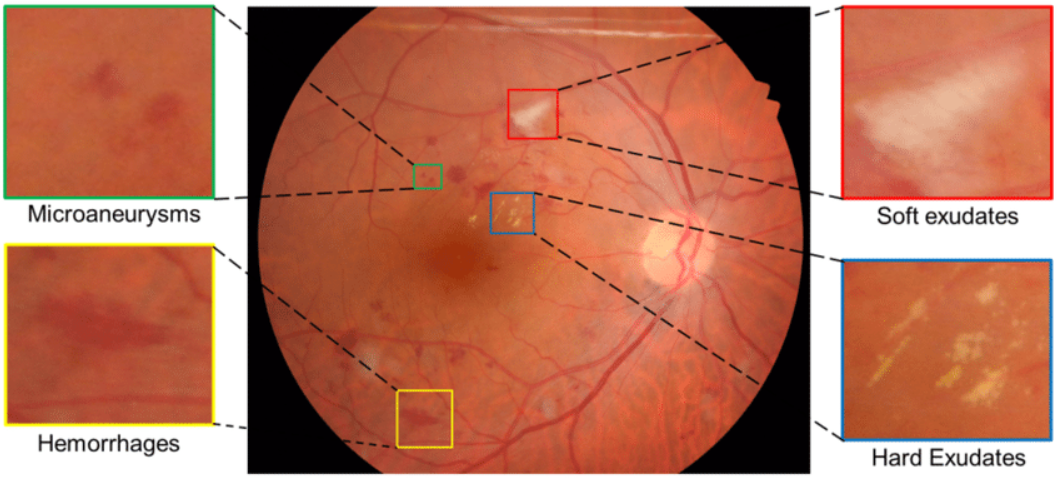}}
    \caption{Retinal Lesions in Diabetic Retinopathy, including MA, HE, SE, and EX. through enhanced sectional Analysis \cite{b4}.}
    \label{fig1}
\end{figure}

Deep learning (DL) is a powerful subset of artificial intelligence (AI) used in computer-aided applications to analyze diabetic retinopathy (DR). DL techniques revolutionize DR analysis by enabling precise, effective, and scalable strategies to detect, diagnose, and monitor the disease \cite{b4}. These advancements have the potential to enhance patient outcomes and reduce the negative impact of vision loss associated with DR. DR-related retinal lesions, like Microaneurysms (MA), Hemorrhages (HE), Exudates (EX), and Soft Exudates (SE) ( see Figure \ref{fig1}), need to be fully automated to reach the primary goal of this work. It also wants to sort fundus pictures by how bad DR and diabetic macular edema are, which is essential for figuring out how fast the disease worsens and what treatments work best. 

\section{Related Works}

The field of Diabetic Retinopathy (DR) analysis has made remarkable progress. This section will explore the latest state-of-the-art methods for detecting DR lesions. We will highlight innovative approaches that use convolutional neural networks (CNNs), deep belief networks (DBNs), and transfer learning to achieve greater accuracy and efficiency.

Retinal image analysis for diabetic retinopathy detection has witnessed significant advancements in various deep-learning approaches. CNN has been essential to the automation of DR grading systems, extracting specific image features through an appropriate weight matrix, and enabling accurate classification of severity levels \cite{b5}. Another automatic DL-based model \cite{b6} incorporates pre-processing, recognition, and detection stages, involving procedures like blood vessel extraction, green channel enhancement, and optic disc removal. This model recognizes DR features using methods such as the RRGS algorithm for hard exudates, LGF, and matched filtering for micro-aneurysms (MA) and hemorrhages (HEM). CNN effectively detects these features based on the extracted data.

Deep Convolutional Neural Networks (DCNNs) have demonstrated prowess in DR detection, with a model developed and tested using a substantial dataset, proposing an ensemble approach involving Inception-V3, Resnet50, Dense-121, Dense-169, and X-ception \cite{b7}. This ensemble of deep CNN models enhances detection accuracy by extracting rich features from retinal images. Deep Belief Networks (DBNs) have been employed in a framework for detecting abnormalities in medical images \cite{b8}. This framework includes stages for feature extraction and detection, with the DBN leveraging self-improved gray wolf optimization for enhanced accuracy. Another study \cite{b9} describes an automatic model incorporating pre-processing, removal of optical disk and blood vessels, abnormality segmentation, and feature extraction and selection, demonstrating effectiveness through meta-heuristic algorithms and optimal feature selection for DBN-based detection.

Transfer learning approaches have been explored, with a deep transfer learning model utilizing an Inception-V3 network \cite{b10}. This model comprises convolutional layers, inception modules, and fully connected layers. It effectively categorizes retinal images and demonstrates the adaptability and accuracy of transfer learning in detecting DR. Additionally, a technique leveraging pre-trained models, including InceptionV3, X-caption, and Inception Resnet-v2, utilizes transfer and ensemble learning for improved performance and was evaluated on the IDRiD test dataset \cite{b11}.
 
\section{Proposed Method}

\begin{figure}[htbp]
    \centerline{\includegraphics[width=\linewidth]{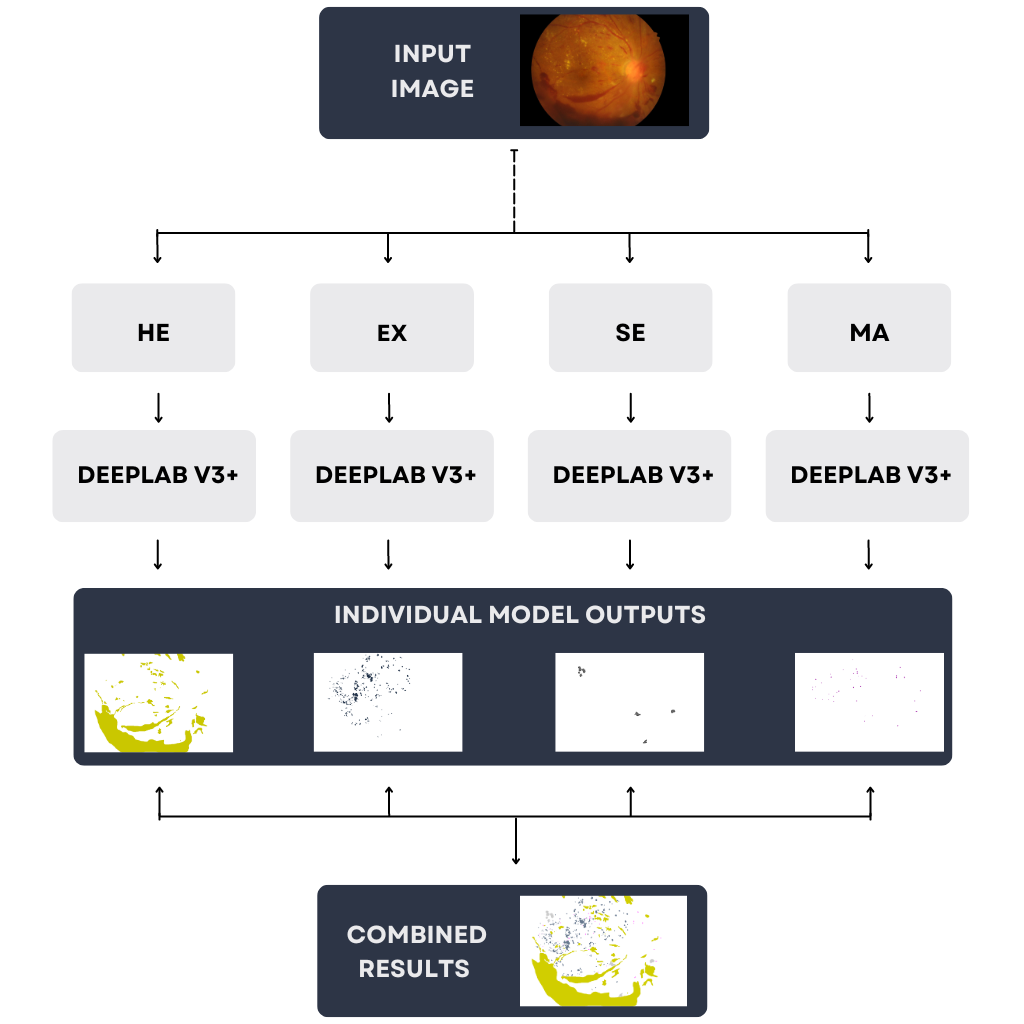}}
    \caption{Overview of the proposed system.}
    \label{fig20}
\end{figure}

To improve the segmentation of diabetic retinopathy classes (MA, HE, EX, and SE), we proposed a binary segmentation approach for each class individually based on the DeepLabV3+ model. This method (see Figure \ref{fig20}) allowed us to fine-tune parameters and enhance accuracy for each category, addressing the complexity of annotations and dataset limitations. After segmenting each class, we combined the results into a single image to improve overall performance, as each class required specific preprocessing parameters.

\begin{figure*}[htbp]
    \centerline{\includegraphics[width=\textwidth]{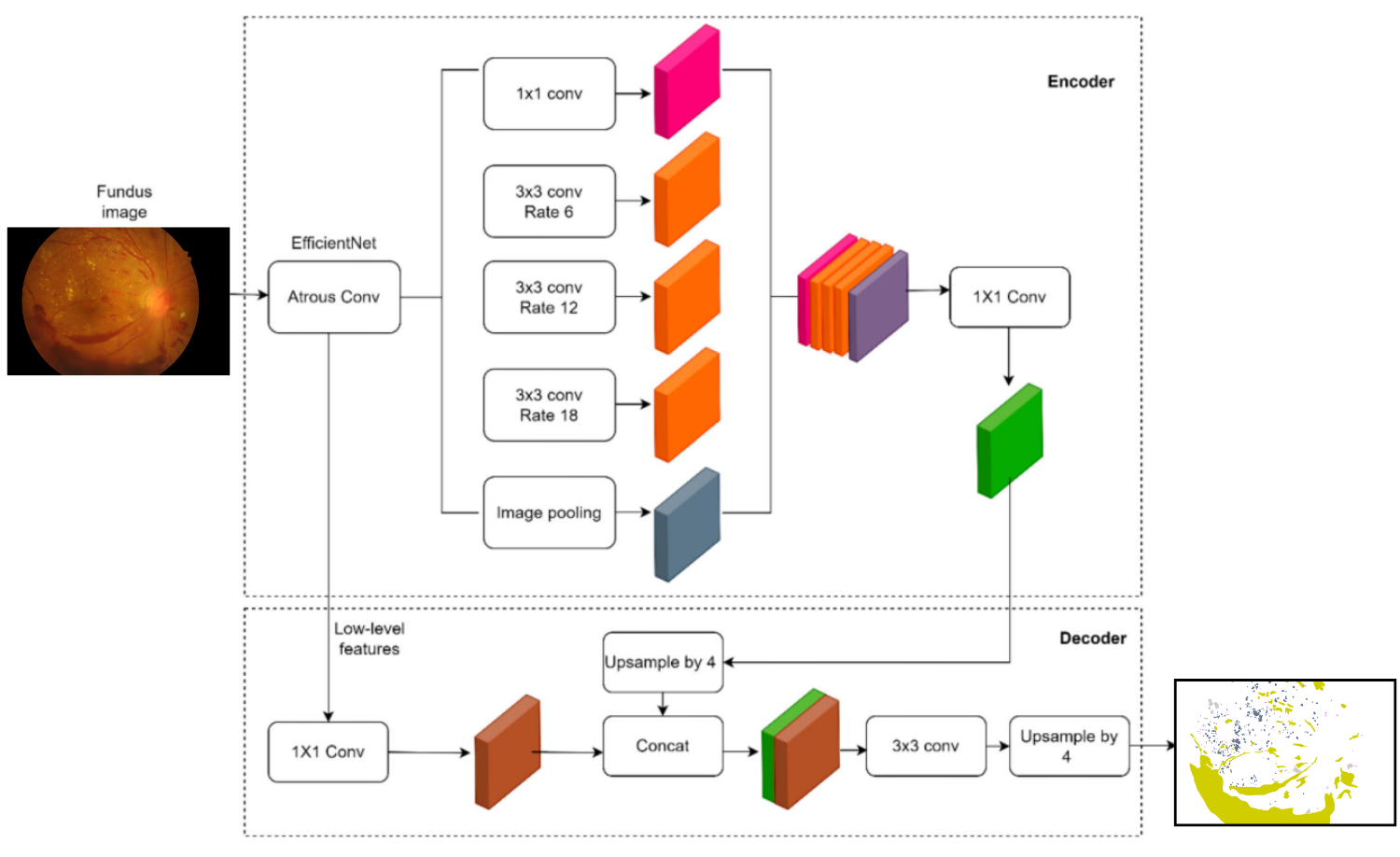}}
    \caption{The DeepLabV3+ architecture overview \cite{b12}.}
    \label{fig22}
\end{figure*}

The cutting-edge deep learning model DeepLabV3+, which is intended for semantic picture segmentation, has proven to perform extraordinarily well in several applications, including medical image analysis. It uses dilated convolution, to record multi-scale contextual data without lowering the feature maps' spatial resolution. In the medical field, precise boundary delineation is crucial for accurate diagnosis and treatment planning, making this skill highly valuable. A sophisticated Atrous Spatial Pyramid Pooling (ASPP) module is integrated into the model to enhance the segmentation of complex anatomical structures by combining information at multiple scales. Through DeepLabV3's robust architecture, we aim to improve the accuracy and efficacy of detecting anomalies in diabetic retinopathy for medical purposes. 
In DeepLabv3+, an encoder-decoder structure is used. The encoder processes contextual information at various scales by using dilated convolutions. Conversely, the decoder improves segmentation results along object boundaries (See Figure \ref{fig22}).

Utilizing configurable decoder and skip connections, a strong backbone network, atrous convolutions, and the ASPP module, DeepLabV3+ provides a powerful and adaptable architecture for semantic segmentation tasks. This combination makes the model highly effective for complex image segmentation tasks, such as those encountered in medical imaging. It enables the model to gather contextual cues and detailed spatial information across various scales. The following are the key components of the DeepLabV3 model:
\begin{itemize}
    \item[--] \textbf{Atrous convolution/Dilated convolution}: maintain a constant stride while increasing the field of view without increasing parameters or computations, resulting in larger output feature maps beneficial for semantic segmentation.

    \item[--] \textbf{Dilated spatial pyramid pooling}: is utilized because the number of valid filter weights decreases as the sampling rate increases. This method is preferred to ensure only valid feature regions are affected, rather than padded zeros.

    \item[--] \textbf{Backbone network}: for feature extraction, it relies on the EfficientNet backbone. This model has strong feature extraction capabilities because it has been pre-trained on huge ImageNet datasets. The ASPP module processes the high-level semantic characteristics that the backbone network derives from the input image.

    \item[--] \textbf{Decoder}: it incorporates a decoder module to improve the segmentation outcomes. The low-resolution feature maps from the ASPP module are upsampled by the decoder. The upsampled maps are then blended with high-resolution data from previous backbone network layers to aid in producing outputs with better segmentation and accurate border delineation.

    \item[--] \textbf{Skip connections}: skip connections are used to merge low-level and high-level features by joining the respective layers in the decoder with the layers from the backbone network. This improves accuracy in segmentation and preserves spatial information around object boundaries.

    \item[--]  \textbf{Sigmoid layer}: the final output is sent via a sigmoid layer to perform binary segmentation. The raw scores are transformed into probabilities for the binary classes (retinal pathology and background) using the sigmoid layer. A probability value, ranging from 0 to 1, is assigned to every pixel in the image, signifying its possibility of being part of the foreground class. This layer uses a threshold to categorize each pixel as belonging to the object or the background, resulting in the final segmented output.

\end{itemize} 
In the next section, we will detail the training and preprocessing steps followed to create our DR anomaly detection solution using the DeepLabV3+ model.

%%%%%%%%%%%%%%%%%%%%%%%%%%%%%%%%%%%%%%
\section{Experiments and results}
\subsection{The used Dataset}
\begin{figure}[htbp]
    \centerline{\includegraphics[width=0.8\linewidth]{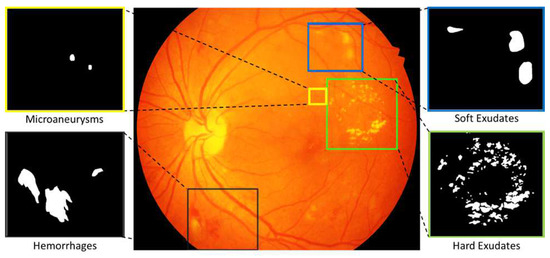}}
    \caption{Enlarged sections highlight MA, SE, HE, and EX \cite{b4}.}
    \label{fig9}
\end{figure}

To train our retinopathy lesion detection and segmentation model, we referred to the Indian Diabetic Retinopathy Image Dataset (IDRID) dataset, representing the Indian population. It is the only dataset with pixel-level annotations of normal retinal structures and typical diabetic retinopathy lesions. It contains 516 retinal images classified as DR and non-DR and offers comprehensive ground truths, including pixel-level annotations for four DR lesions (EX, HE, MA, and SE). Figure \ref{fig9} provides examples of an image from the dataset and its corresponding ground truths.

\begin{figure}[htbp]
    \centerline{\includegraphics[width=\linewidth]{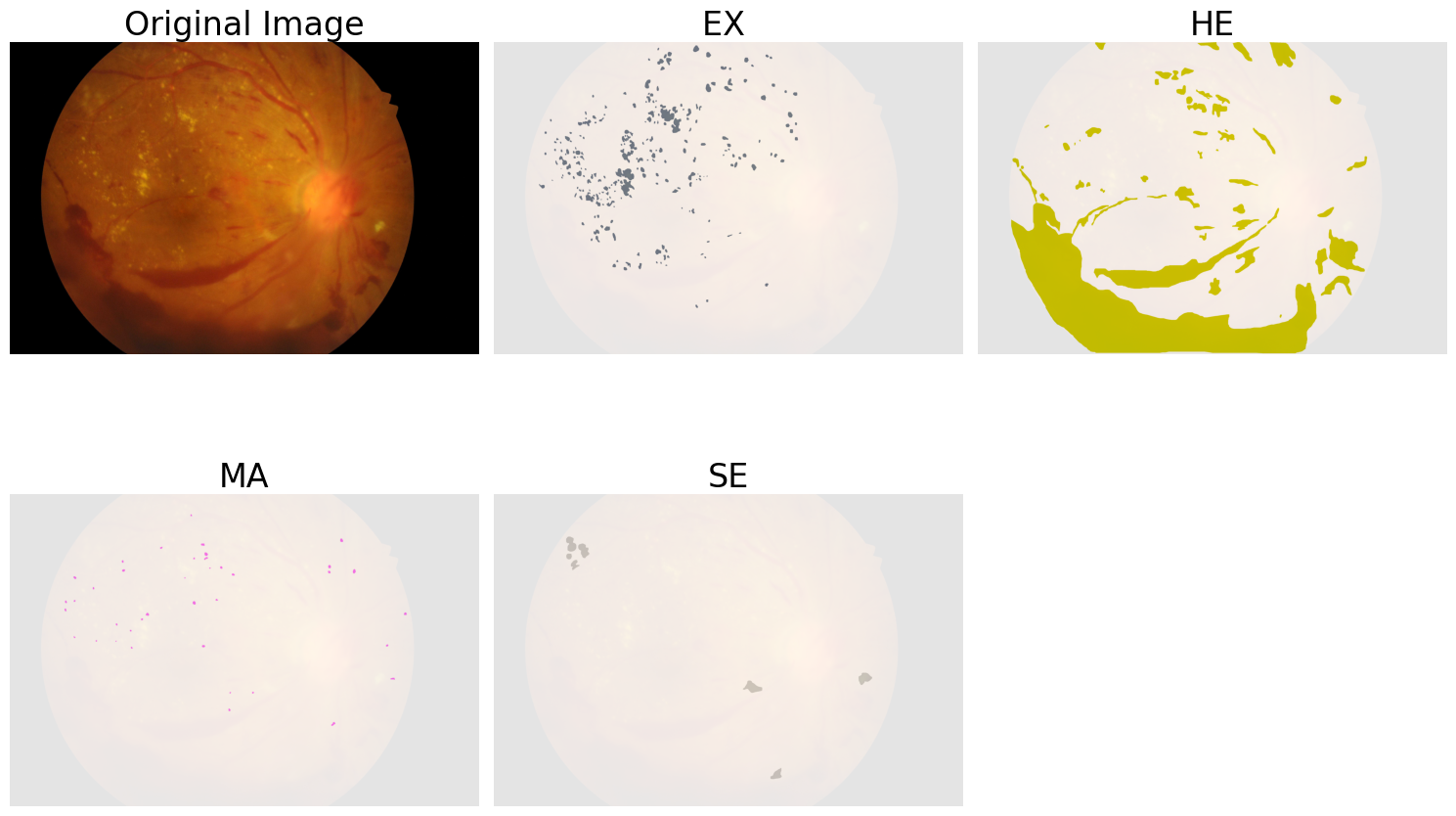}}
    \caption{Original image and masks.}
    \label{fig18}
\end{figure}

Figure \ref{fig18} shows an example image from the dataset, combined with its ground truth for better visibility. This demonstrates the presence of multiple pathological classes. Each image may contain one or more pathological classes (EX, MA, SE, and HE). The dataset captures the complexity and variability of retinal pathology through images containing multiple classes.

\subsection{Data Preprocessing}

\begin{figure}[htbp]
    \centerline{\includegraphics[width=\linewidth]{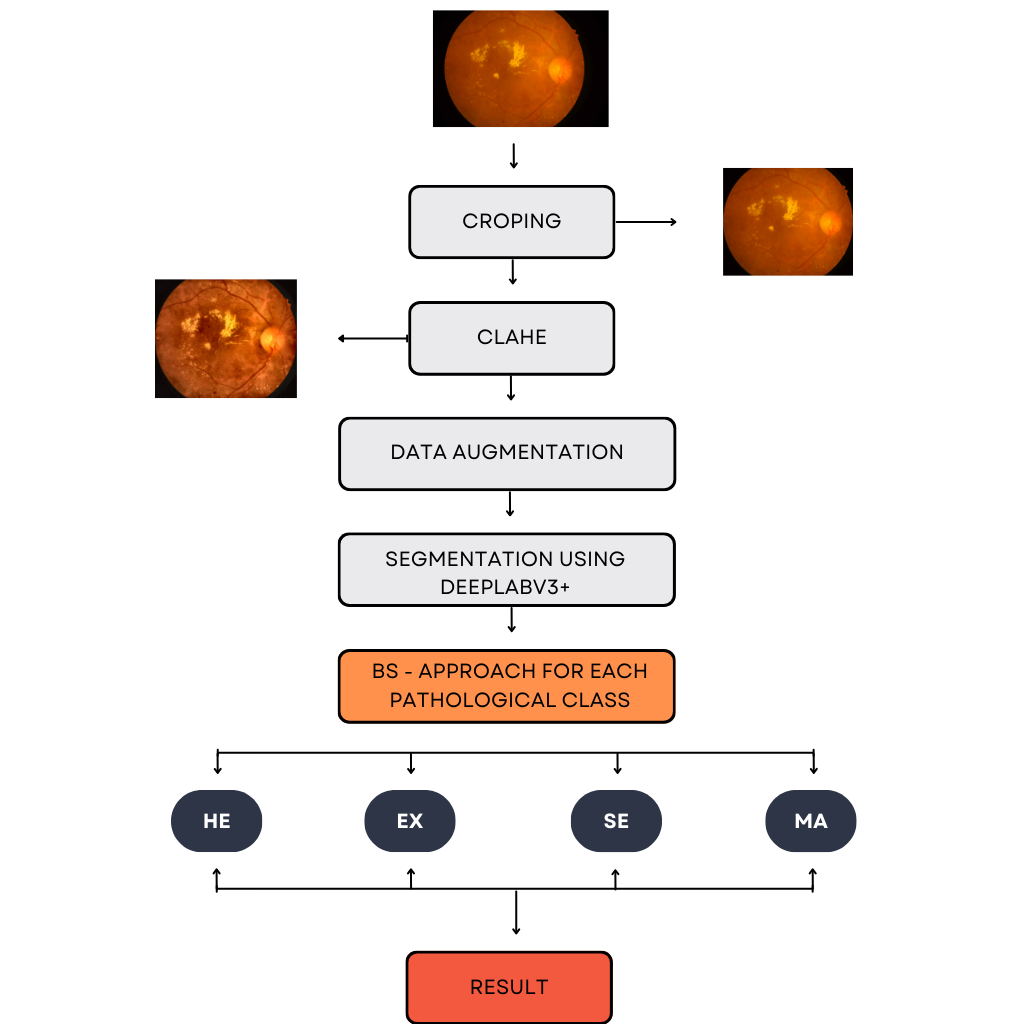}}
    \caption{Overview of the preprocessing stage. \textit{*BS: binary segmentation}}
    \label{fig21}
\end{figure}

Data preprocessing is essential for improving the quality and contrast of retinal fundus images. Figure \ref{fig21} represents the key steps applied in our preprocessing step. We began by cropping the image to delete undesired elements, allowing for a focused analysis of the fundus. Furthermore, we utilized two crucial data augmentation techniques: rotation and horizontal flipping. This increased dataset variability while preserving image integrity. We then used CLAHE for contrast enhancement, focusing on the L channel. 

In the remainder of this section, we will provide more details about the preprocessing steps.

\subsubsection{Croping} 
\begin{figure}[htbp]
    \centerline{\includegraphics[width=0.4\textwidth]{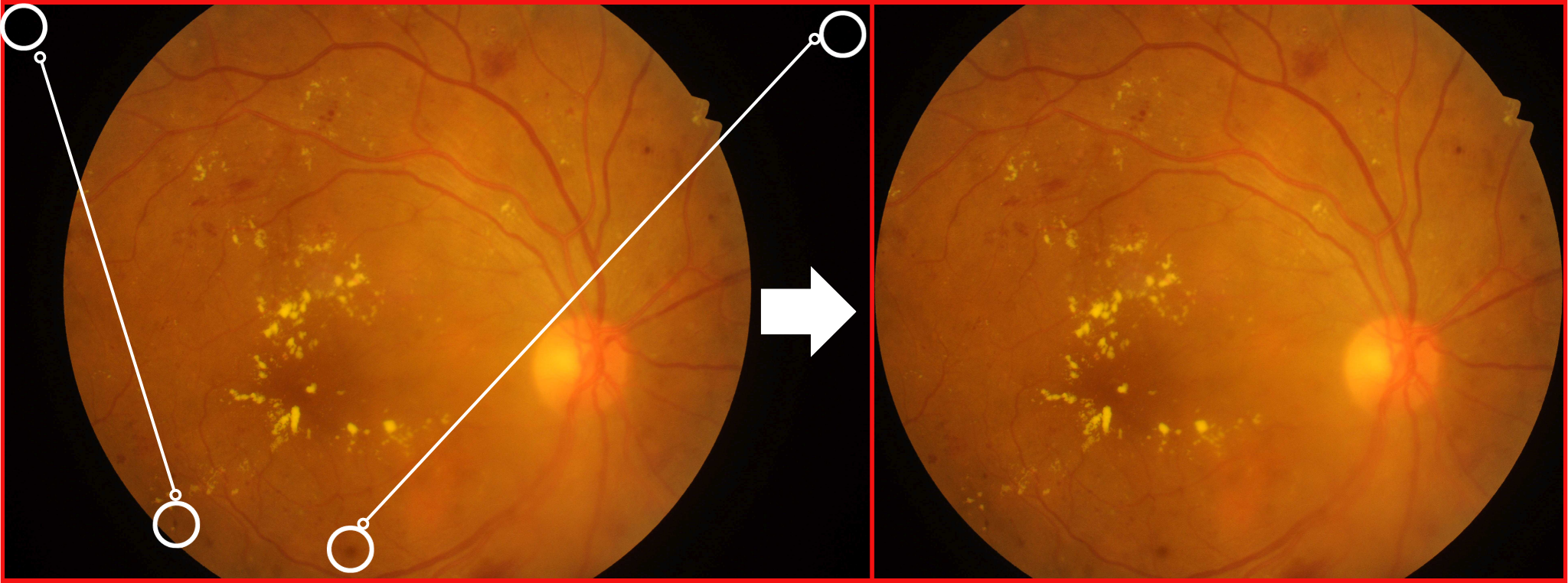}}
    \caption{Enhancing lesion discrimination by removing background similarities.}
    \label{fig7}
\end{figure}

we utilized an advanced image cropping algorithm to enhance the input data. This method aimed to eliminate unnecessary parts of the background, particularly the image corners, as determined through extensive testing (see Figure \ref{fig7}).
The gray intensities of the pixels that make up the ground truth and the eye region were analyzed, and we found that some of the pixels had very similar gray intensities (see figure \ref{fig7}). This might potentially interfere with the model's ability to learn and cause confusion in the final results, which would reduce precision. We employed a cropping method that selectively eliminated undesired elements while preserving the essential circular structure of the eye in the DR image. This was necessary as the background exhibited some subtle color noise. The application of this cropping technique yielded excellent results, significantly enhancing the accuracy and relevance of the cropped images for subsequent analysis.

\subsubsection{Used CLAHE} 
\begin{figure}[htbp]
    \centerline{\includegraphics[width=0.35\textwidth]{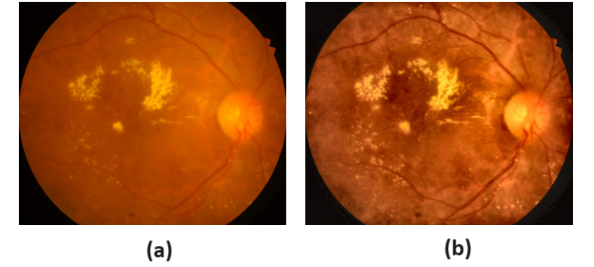}}
    \caption{An image from the IDRiD dataset before (a) and after (b) applying CLAHE.}
    \label{fig2}
\end{figure}

a powerful tool for enhancing the quality of images in various applications, especially in medical imaging. It is a sophisticated technique that improves the local contrast of an image (see Figure \ref{fig2}), particularly in areas with varying illumination \cite{b13}.

\begin{figure}[htbp]
    \centerline{\includegraphics[width=0.5\textwidth]{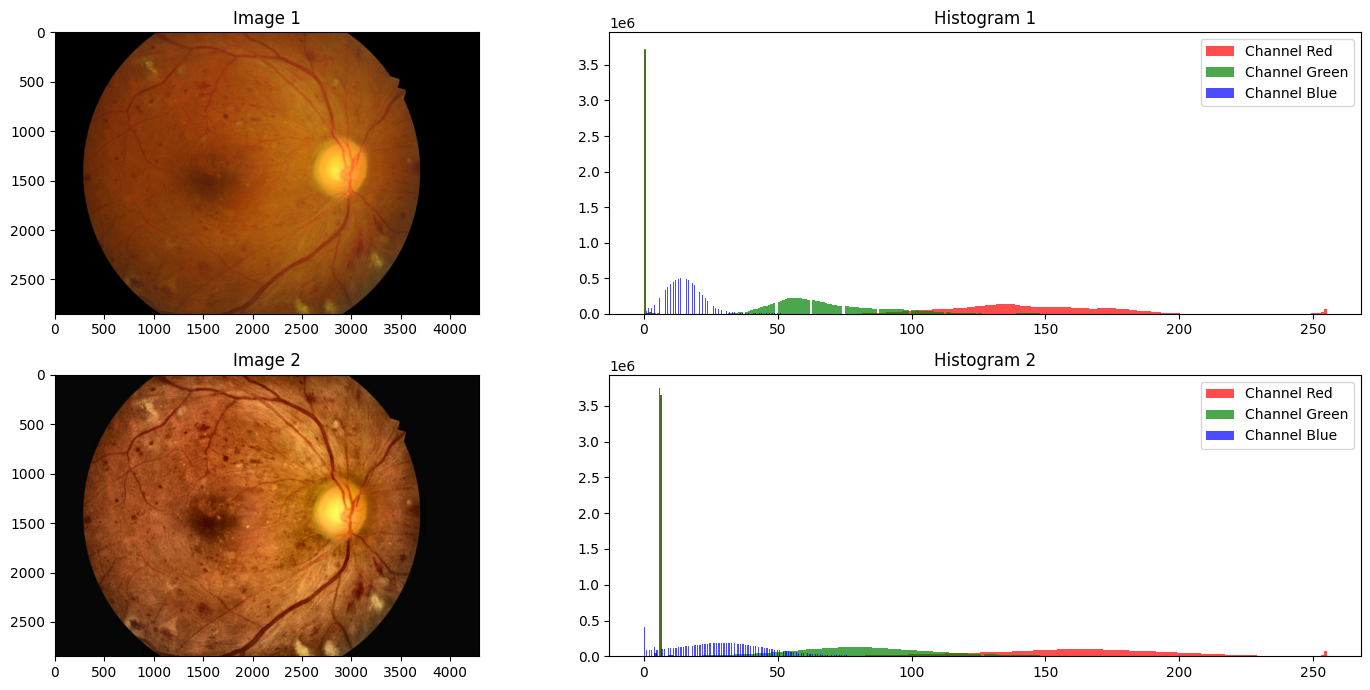}}
    \caption{Histogram comparison: original (image 1) vs CLAHE enhanced image (image 2).}
    \label{fig3}
\end{figure}

In the preprocessing step, we applied CLAHE to the $LAB$ color space, focusing on the $L$ channel, where $A$ indicates the red/green coordinate, $B$ signifies the yellow/blue coordinate and $L$ indicates lightness (See Figure \ref{fig3}). 

The process involved splitting the $LAB$ image into its components ($L$, $A$, and $B$), applying CLAHE to the $L$ channel, and merging it with the $A$ and $B$ channels. This method effectively improved the visibility of areas with irregular illumination. The histograms visually represent the distribution of pixel intensities in the images both before and after the CLAHE application. This method aims to increase local contrast and make overall image details more visible, which enhances our proposed method's performance.

\subsubsection{Data splitting}
the challenging dataset consists of limited annotations: 81 images/masks for EX and MA lesions, 80 images/masks for HE lesions, and 40 images/masks for SE lesions. To ensure unbiased evaluation, we first split the original images into training (70\%), validation (20\%), and testing (10\%) sets before augmentation. This prevents the model from seeing the validation and testing images during training.

\subsubsection{Data augmentation} After experimenting with different data augmentation methods, we discovered that when working with carefully annotated medical images, applying too many techniques can result in losing important information. For example, aggressive transformations such as large-scale cropping or excessive warping can distort anatomical structures, making it challenging for the model to learn relevant features. So, we focused on two essential methods: rotation (with a range of -10 to 10 degrees) and horizontal random flipping (50\% probability), which help retain information and pixel uniformity. Due to the dataset limitation, we created nine augmented samples per original image. This number was chosen to preserve the integrity of the annotations, considering their complexity in medical imaging. Table \ref{tab0} presents the image distributions of each class before and after the augmentation. 

\begin{table}[htbp]
\caption{ Image distributions of each class before and after the augmentation.}
\label{tab0}
\centering
\begin{tabular}{|l|l|l|l|}
\hline
\textbf{Lesion Type} & \textbf{Split} & \textbf{Original Data} & \textbf{Augmented Data} \\ \hline
\multirow{3}{*}{EX and MA} & Train       & 57            & 513            \\ \cline{2-4} 
                           & Validation  & 16            & -              \\ \cline{2-4} 
                           & Test        & 8             & -              \\ \hline
\multirow{3}{*}{HE}        & Train       & 56            & 504            \\ \cline{2-4} 
                           & Validation  & 16            & -              \\ \cline{2-4} 
                           & Test        & 8             & -              \\ \hline
\multirow{3}{*}{SE}        & Train       & 28            & 252            \\ \cline{2-4} 
                           & Validation  & 8             & -              \\ \cline{2-4} 
                           & Test        & 4             & -              \\ \hline
\end{tabular}
\end{table}

\subsection{Training}

We trained the DeepLabV3+ model on the train dataset using the hyperparameters presented in Table \ref{tab1}. 

\begin{table}[htbp]
    \centering
    \caption{Parameters and Hyperparameters used in the Training Process}
    \label{tab1}
    \begin{tabular}{|p{3.5cm}|p{4.5cm}|}
    \hline
    \textbf{Parameter/Hyperparameter} & \textbf{Value} \\
    \hline
    Image Size & 512 \\
    \hline
    Batch Size & 4 \\
    \hline
    Epoch  & 30 \\
    \hline
    Number of Classes & 1 (1 for the DR class and 0 for the background) \\
    \hline
    Loss Function & BinaryCrossentropy \\
    \hline
    activation function & sigmoid \\
    \hline
    Optimizer & Adam \\
    \hline
    Learning Rate & 0.0001 \\
    \hline
    Early Stopping Monitor & val\_loss \\
    \hline
    Early Stopping Patience & 5 \\
    \hline
    \end{tabular}

\end{table}

\begin{figure}[htbp]
    \centerline{\includegraphics[width=0.5\textwidth]{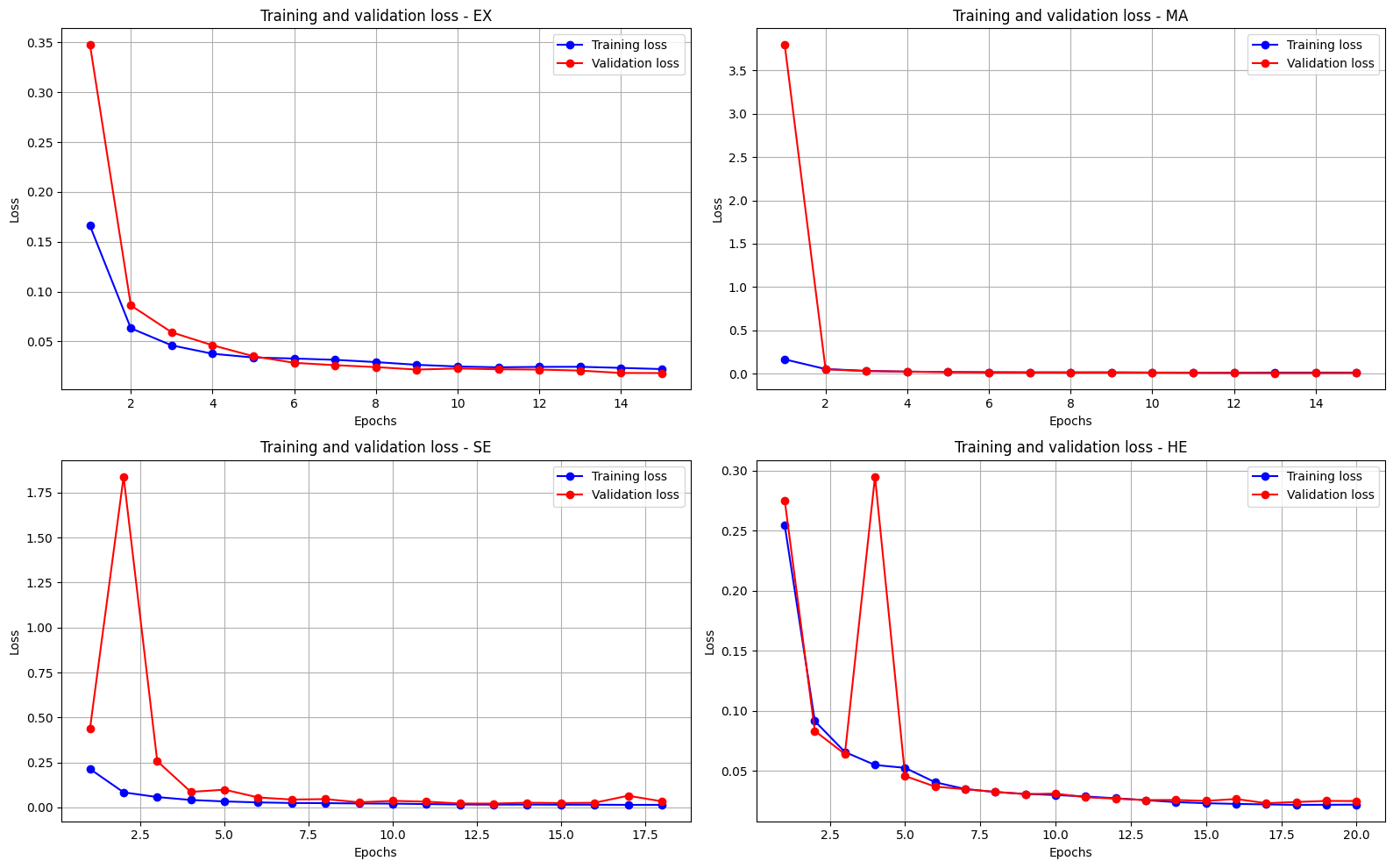}}
    \caption{Training and validation loss.}
    \label{fig23}
\end{figure}

The training and validation loss curves for the EX, MA, SE, and HE classes illustrate the effectiveness and robustness of our models across various datasets (See Figure \ref{fig23}). All error curves decrease and converge to zero, with minimal gaps between the training and validation curves indicating a very good fit. These curves affirm the models' robustness and their ability to generalize well across various datasets, showcasing the effectiveness of our training approach.

% \color{blue} Pourquoi tu n'as pas ajouté les courbes des autres métriques? on a encore de l'éspace.\color{black}

% Meher: here are all the metrics of the deeplabv3+! 

%------------------éliminer ça ou non?
%The MA class exhibits a rapid decrease in losses with minimal divergence between training and validation, showcasing the model's robustness and ability to generalize exceptionally well. Although the SE class initially displays fluctuations in validation loss, the model stabilizes over time, indicating it eventually overcomes complexities and noise in the data to generalize effectively. The HE class, despite a brief spike in validation loss, quickly recovers and achieves a stable convergence, further demonstrating the model's resilience and strong learning capabilities. Overall, the EX and MA classes highlight our best work, with low and stable losses, underscoring the models' high performance and reliability. These results affirm the models' robustness and their ability to generalize well across various datasets, showcasing the effectiveness of our training approach.

% Meher: Yes, we can delete this! 

\color{black}

\subsection{Evaluation and test}

Our segmentation results, meticulously evaluated on the test dataset through a range of metrics Accuracy, Specificity, Sensitivity, F1 score, and IoU (see formulas in Tables \ref{tab2},\ref{tab3}), underscore the effectiveness of our method in addressing each pathological class. These results highlight the importance of our proposed method in ensuring precise and reliable segmentation across various pathological categories (refer to Table \ref{tab4}).

\begin{table}[htbp]
\caption{Evaluation metrics and their formulas.}
\label{tab2}
\centering
\renewcommand{\arraystretch}{1.5} % Adjusts the vertical padding
\begin{tabular}{|p{2cm}|p{3.5cm}|}
\hline
\textbf{Metric} & \textbf{Formula} \\ \hline
Accuracy        & $= \frac{TP + TN}{TP + TN + FP + FN}$ \\ \hline
Specificity     & $ = \frac{TN}{TN + FP}$ \\ \hline
Sensitivity     & $ = \frac{TP}{TP + FN}$ \\ \hline
F1 Score        & $ = 2 \cdot \frac{\text{Precision} \cdot \text{Recall}}{\text{Precision} + \text{Recall}}$ \\ \hline
IoU & $= \frac{\text{Area of Overlap}}{\text{Area of Union}}$ \\ \hline
MAE & $ = \frac{1}{n} \sum_{i=1}^{n} |y_i - \hat{y_i}|$ \\ \hline
MSE & $= \frac{1}{n} \sum_{i=1}^{n} (y_i - \hat{y_i})^2$ \\ \hline
\end{tabular}
\end{table}

\begin{table}[htbp]
\caption{Definitions of variables used in the formulas.}
\label{tab3}
\centering
\renewcommand{\arraystretch}{1.5}
\begin{tabular}{|p{1.1cm}|p{6.8cm}|}
\hline
\textbf{Variable} & \textbf{Definition} \\ \hline
$TP$ & True Positives: The number of correctly predicted positive cases. \\ \hline
$TN$ & True Negatives: The number of correctly predicted negative cases. \\ \hline
$FP$ & False Positives: The number of incorrectly predicted positive cases. \\ \hline
$FN$ & False Negatives: The number of incorrectly predicted negative cases. \\ \hline
$y_i$ & The actual value for the $i$-th instance. \\ \hline
$\hat{y_i}$ & The predicted value for the $i$-th instance. \\ \hline
$n$ & The total number of instances. \\ \hline
\end{tabular}
\end{table}

\begin{table}[htbp]
\caption{Evaluation metrics scores for EX, HE, MA, and SE classes on the test dataset.}
\label{tab4}
\renewcommand{\arraystretch}{1.5}
\begin{center}
\begin{tabular}{|c|c|c|c|c|}
\hline
\textbf{Metric} & \textbf{EX} & \textbf{HE} & \textbf{MA} & \textbf{SE} \\
\hline
Accuracy & 0.99391 & 0.99272 & 0.99789 & 0.99774 \\
\hline
Specificity & 0.99778 & 0.99866 & 0.99886 & 0.99994 \\
\hline
Sensitivity & 0.99606 & 0.99397 & 0.99903 & 0.99780 \\
\hline
F1 Score & 0.99692 & 0.99630 & 0.99894 & 0.99887 \\
\hline
IoU & 0.99386 & 0.99265 & 0.99789 & 0.99774 \\
\hline

\end{tabular}
\end{center}
\end{table}

\begin{figure}[htbp] \centerline{\includegraphics[width=0.5\textwidth]{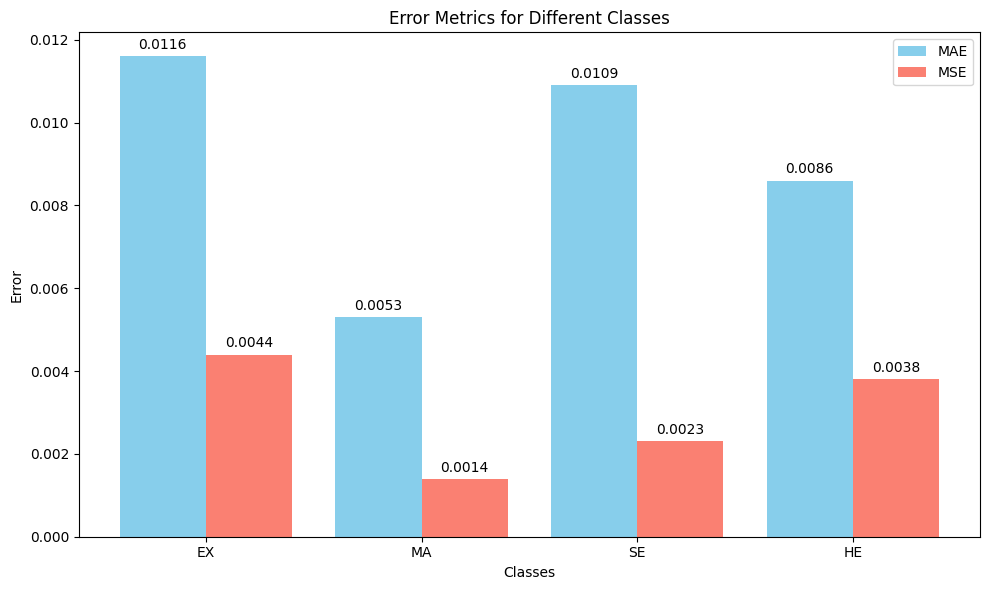}}
    \caption{ Error metrics MSE \& MAE .}
    \label{fig24}
\end{figure}

To evaluate and better analyze the trained model's performances, we also calculated the Mean Average Error (MAE) and the Mean Squared Error (MSE) on the test dataset for the four classes. As shown in Figure \ref{fig24}, the MA class achieved the lowest MSE of 0.0014, showcasing exceptional model performance. Overall, the trained model demonstrates strong performance across all classes, with a particularly outstanding performance in the MA class. This underscores the effectiveness of our training approach and the model's ability to generalize well across the test dataset samples.

%-------------------- Je ne sais pas si on doir 

%mettre ces détails ou pas? à madame Afef

%The EX class also showed good performance with accurate predictions and minimal deviation. The SE class performed well, with slightly higher errors than the MA class but still within an acceptable range. The HE class exhibited robust performance with reliable predictions. 

% Meher: we can delete this I think! 

\color{black}

\begin{figure}[htbp]
    \centerline{\includegraphics[width=0.5\textwidth]{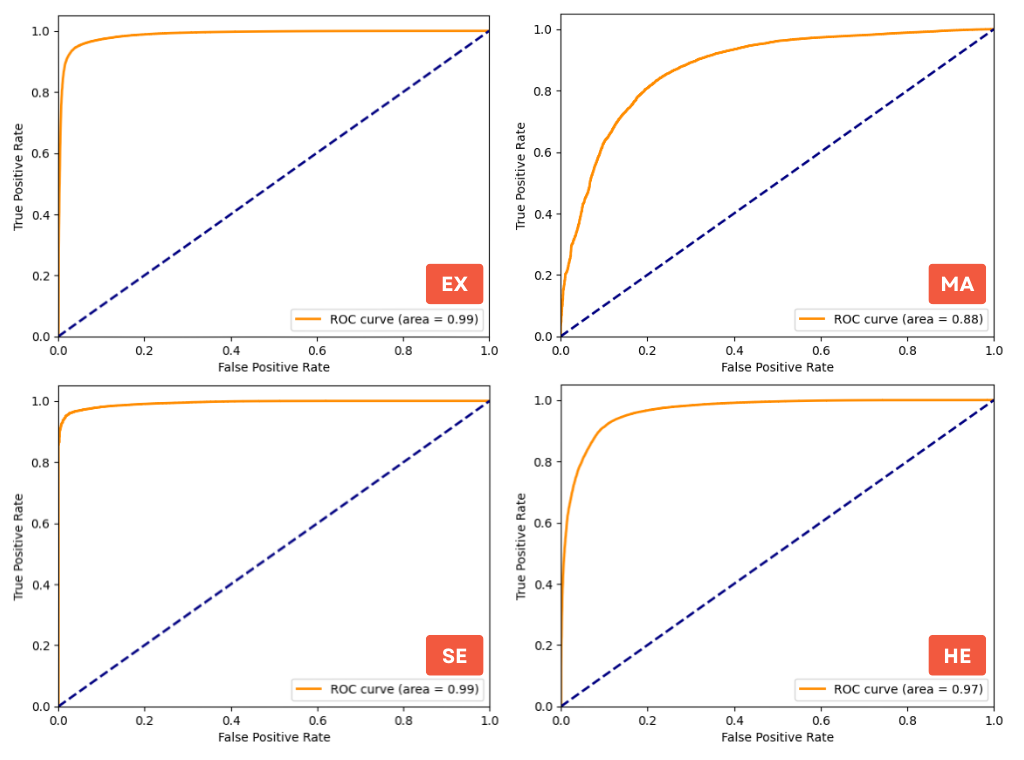}}
    \caption{ ROC curve of all classes .}
    \label{fig25}
\end{figure}

We also plotted the Receiver Operating Characteristic (ROC) curves to demonstrate the model's performance in distinguishing between the four classes (see Figure \ref{fig25}). The model performs well in all classes, with the SE and EX classifications showing very strong performance. These strong AUC values demonstrate the model's ability to correctly categorize data items into various classes. 

% \color{blue} Pourquoi dans la section d'avant, on a eu la meilleure performance est celle de la classe MA et ici on dit que les meilleures performances sont ceux des classes SE et EX. C'est contradictoire! il est preférable de parler de la bonne performance en globale pour les 4 classes. \color{black} 

% Meher: Done! 

\begin{figure}[htbp]
    \centerline{\includegraphics[width=0.5\textwidth]{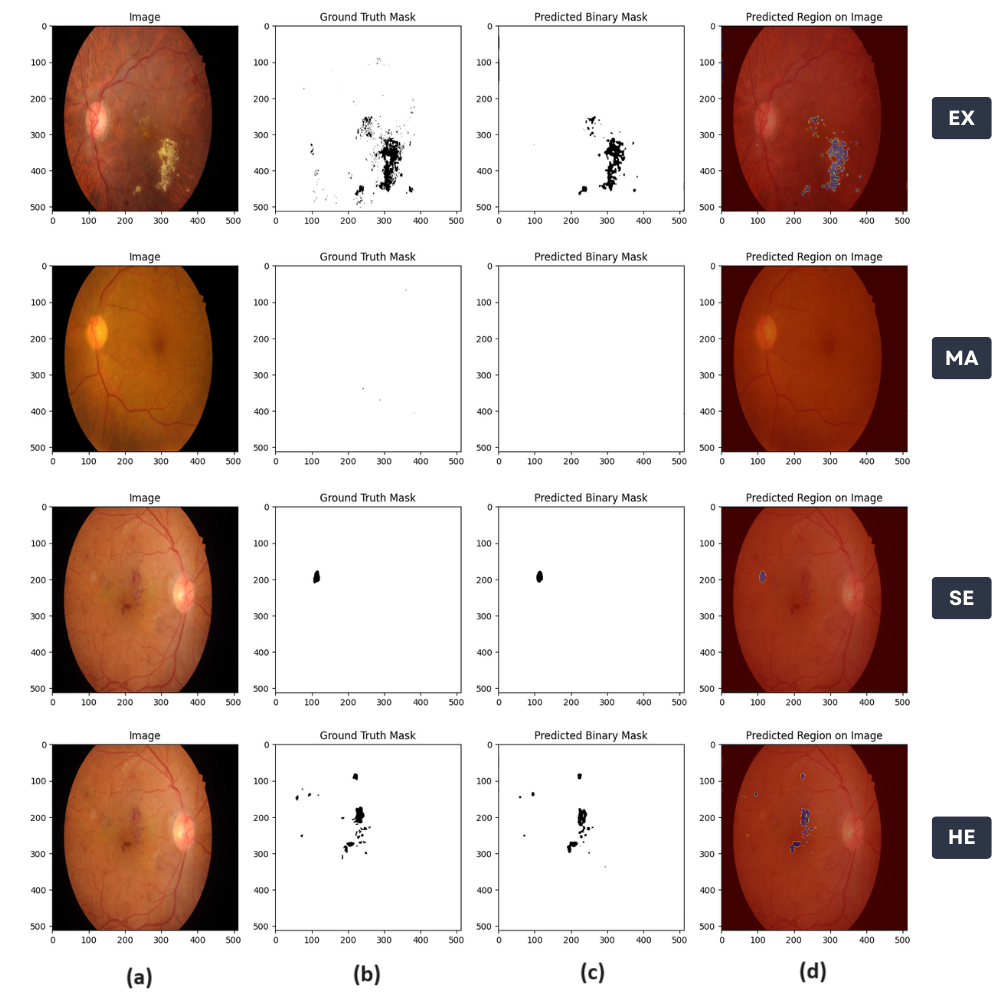}}
    \caption{ Examples of inferences: (a) test images, (b) ground truth masks, (c) model results, and (d) detected regions combined with original images.}
    \label{fig26}
\end{figure}

Figure \ref{fig26} presents examples of the trained model inferences on images from the test dataset. As depicted in the figure, the model predictions closely match the ground truth, visually displaying masks for the EX, MA, SE, and HE classes. These results are impressive and demonstrate the model's ability to accurately identify pathological elements within retinal images.

In our study, we went beyond just comparing different types of lesions. We examined the challenges associated with each lesion class and refined our methods through multiple analyses. As a result, we achieved outstanding accuracy rates of over 99.4\% for all classes, surpassing the performance of state-of-the-art methods (refer to Table \ref{Tab5}). These results demonstrate that our DeepLabV3+ model-based method is effective and constitutes a major advancement in the segmentation of DR lesions.

\section{Conclusion and Future Work}

This research has significantly addressed critical challenges in diabetic retinopathy lesion detection and segmentation. By focusing on lesion separation and improving annotations for hard exudates, hemorrhages, microaneurysms, and soft exudates classes, we trained the DeepLabV3+ model on the IDRID dataset and achieved a state-of-the-art accuracy of 99\%. We achieved high accuracy by combining meticulous data preprocessing and well-studied augmentation techniques. The key factors enabling us to overcome these obstacles were careful fine-tuning parameters, refined cropping techniques, and customized histogram modifications. Looking forward, further research could enhance the significance of this proposed approach. Additional exploration into advanced preprocessing techniques adapted to specific lesion features might further improve segmentation accuracy.

\begin{table*}[htbp]
\caption{Comparison with the state-of-the-art methods on the IDRiD database.}
\label{Tab5}
\centering
\renewcommand{\arraystretch}{1.5} % Adjusts the vertical padding
\begin{center}
\begin{tabular}{|p{1cm}|p{1cm}|p{7cm}|p{4.2cm}|p{2.8cm}|}
\hline
\textbf{Method} & \textbf{Year} & \textbf{Methodology} & \textbf{Segmentation Techniques} & \textbf{Results} \\
\hline
\cite{b14} & 2018 & U-Net, HEDNet, HEDNet+cGAN, Conditional Generative Adversarial Network (cGAN), PatchGAN, VGG16 Weighted Binary Cross-Entropy, Loss, CLAHE, Bilateral Filter & U-Net, HEDNet, HEDNet+cGAN  & Precision = 84.05\% \\
\hline
\cite{b15} & 2018 & Deep-CNN, Binary Cross Entropy, VGG16 & Deep-CNN  & IOU = 85.72\% \\
\hline
\cite{b16} & 2019 & Deep CNN, DeepLabV3, Segnet, Conditional Random Field (CRF) & DeepLabV3, Segnet  & ACC = 88\% \\
\hline
\cite{b17} & 2020 & U-Net, ResNet34, Initialized to Convolution NN Resize (ICNR) & U-Net  & ACC = 99.88\%\\
\hline
\cite{b18} & 2021 & CNN U-Net, AlexNet, VGGNet, Green Channel, Adam Optimizer & CNN U-Net  & ACC = 98.68\% \\
\hline
\cite{b19} & 2021 & Adaptive Active Contour, Otsu Thresholding, Morphological Operation, Median Filtering, Open-Close Watershed Transform, GLCM, ROI, LTP & Adaptive Active Contour, Watershed Transform, Otsu Thresholding  & ACC = 60\% \\
\hline
\cite{b20} & 2021 & EAD-Net, U-Net, CAM, PAM & EAD-Net, U-Net & ACC = 78\% \\
\hline
\cite{b21} & 2023 & Circular Hough Transform, Morphological Operations, Average Histogram, Contrast Enhancement, CCA & Circular Hough Transform  & SF = 96.80\% \\
\hline
\textbf{Our work} & \textbf{2023} & \textbf{Binary segmentation for each pathological class: cropping, data augmentation, CLAHE, Binary Cross Entropy, Atrous Convolution, Dilated Spatial Pyramid Pooling, Adam Optimizer } & \textbf{Deeplabv3+} & \textbf{ACC = 99\%, IOU = 99\%, F1 Score = 99\%, SN = 99\%, SF = 99\%} \\
\hline 
\end{tabular}
\end{center}
\end{table*}

\end{document}